\documentclass[10pt,twocolumn,letterpaper]{article}

\usepackage[pagenumbers]{cvpr} 

\usepackage[dvipsnames]{xcolor}

\usepackage{bbding}
\usepackage{longtable}
\usepackage{graphicx}
\usepackage{multirow}
\usepackage{bm}
\usepackage{soul}
\usepackage{url}
\usepackage{dsfont}
\usepackage{algorithm}
\usepackage{algorithmicx}
\usepackage{algpseudocode}
\usepackage{mathtools}
\newcommand{\minbox}[2]{%
  \mathmakebox[\ifdim#1<\width\width\else#1\fi]{#2}}

\algnewcommand{\Local}{\State\textbf{local variables: }}
\usepackage{makecell}
\usepackage{wrapfig}
\usepackage{vcell}
\usepackage{array}
\usepackage[switch]{lineno}
\usepackage{times}
\usepackage{epsfig}
\usepackage{amsmath}
\usepackage{amssymb}
\usepackage{pifont}%

\usepackage{multirow}
\usepackage{bbding}
\usepackage{microtype}      
\usepackage{xcolor}         
\usepackage{relsize}
\usepackage{xspace}
\usepackage{booktabs}

\newtheorem{proposition}{Proposition}

\definecolor{cvprblue}{rgb}{0.21,0.49,0.74}
\usepackage[pagebackref,breaklinks,colorlinks,citecolor=cvprblue]{hyperref}

\usepackage[capitalize]{cleveref}
\crefname{section}{Sec.}{Secs.}
\Crefname{section}{Section}{Sections}
\Crefname{table}{Table}{Tables}
\crefname{table}{Tab.}{Tabs.}

\def\ie{\emph{i.e}\onedot} 
 
\def\etc{\emph{etc}\onedot} 
 
\def\etal{\emph{et al}\onedot}

\newcommand{\cmark}{\ding{51}}%
\newcommand{\xmark}{\ding{55}}%

\def\paperID{3744} 

\def\confName{CVPR}
\def\confYear{2024}

\begin{document}

\title{
 Learning Degradation-Independent Representations for Camera ISP Pipelines} 

\author{Yanhui Guo\\
McMaster University, Canada\\
{\tt\small guoy143@mcmaster.ca}
\and
Fangzhou Luo\\
McMaster University, Canada\\
{\tt\small luof1@mcmaster.ca}
\and
Xiaolin Wu\\
McMaster University, Canada\\
{\tt\small xwu@mcmaster.ca}
}

\maketitle

\begin{abstract}

Image signal processing (ISP) pipeline plays a fundamental role in digital cameras, which converts raw Bayer sensor data to RGB images. However, ISP-generated images usually suffer from imperfections due to the compounded degradations that stem from sensor noises, demosaicing noises, compression artifacts, and possibly adverse effects of erroneous ISP hyperparameter settings such as ISO and gamma values. 
In a general sense, these ISP imperfections can be considered as degradations. The highly complex mechanisms of ISP degradations, some of which are even unknown, pose great challenges to the generalization capability of deep neural networks (DNN) for image restoration and to their adaptability to downstream tasks.
To tackle the issues, we propose a novel DNN approach to learn degradation-independent representations (DiR) through the refinement of a self-supervised learned baseline representation. The proposed DiR learning technique has remarkable domain generalization capability 
and consequently, it outperforms state-of-the-art methods across various downstream tasks, including blind image restoration, object detection, and instance segmentation, as verified in our experiments. 

\end{abstract}

\section{Introduction} 

In our information era, digital cameras have become ubiquitous and indispensable in all walks of socioeconomic life. 
They are arguably the most common type of sensory device for both humans and their intelligent agents to sense the environment.  
The applications of digital imaging fall into two categories depending on their end purposes: human vision oriented and machine vision oriented.  Examples of the first category include smartphones, social media, internet videos, \etc, while robotics, automation, and medical imaging are examples of the second category.  

For human vision oriented applications, the image quality metrics should correspond to human visual perception, because the images are primarily intended to please our eyes.  On the other hand, for machine vision oriented applications \cite{jiao2022fine,redmon2018yolov3,xie2021segformer,jiao2023}, the image quality requirements may not completely align with those of human perception.  In the latter case, what matters most is the preservation of semantically critical image features that are needed to solve computer vision problems of recognition and classification nature. 

For either human viewing or downstream computer vision tasks, the image acquisition process of digital cameras is the same.  All images are generated by a hardware image signal processing (ISP) pipeline that transforms raw Bayer sensor data into RGB images.  The ISP pipeline consists of a set of cascaded steps, including denoising, color demosaicking, and compression.  Each of the ISP steps can cause a level of degradation in image quality.  At the beginning of the ISP pipeline, there are sensor noises, particularly in dark environments with a low signal-to-noise ratio.  The noisy RAW data are then demosaiced to produce the corresponding RGB images. This demosaicking process introduces interpolation noises of its own. Finally, the RGB image is compressed for bandwidth efficiency, which adds extra compression noises on top of sensor noises and demosaicking noises. The resulting compound effects of these cascaded noises are very difficult to model precisely \cite{guo2022data}.  The underlying complex image quality degradation mechanism prevents the synthesis of clean and degraded image pairs for supervised learning.

Yet, there is another type of ISP degradation that exerts adverse effects on the applications of ISP-generated images in downstream tasks. The camera ISP involves a number of adjustable hyperparameters such as gamma correction, ISO, gain control, and white balance. Their optimal settings 
may vary from application to application, \ie, there exists no one-fits-all solution. Most cameras have a default hyperparameter setting that is tailored to perceptual quality rather than optimized for a specific computer vision task. 
We consider the deviations of ISP hyperparameters from their optimal settings, in a general sense, to be degradations as well.  This second type of ISP degradation can lead to underperformances or even outright failures of downstream computer vision algorithms \cite{isp_opt_cvpr20,qin2022attention}. 
Like the first type of signal-level degradations, the second type of semantic-level degradations is extremely difficult, if not impossible, to model and synthesize, not only because the space of ISP hyperparameters is of very high dimensions but also because the two types of ISP degradations are intertwined.

\begin{figure}[htbp]
  \centering
  \includegraphics[width=0.4\textwidth]{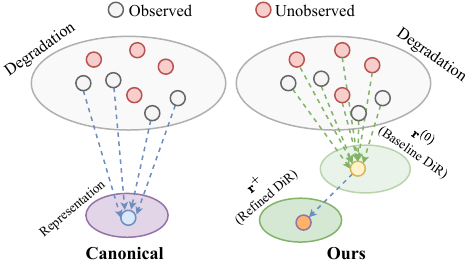}
  \caption{Comparison of the canonical learning paradigm and our DiR learning technique.}
  \label{fig_DI_motivation}
  \vspace{-10pt}
\end{figure}

As discussed above, it is either intractable or highly expensive to obtain the ISP-degraded and the corresponding latent degradation-free (ground truth) image pairs.  Moreover, exhausting all ISP degradation variants in a brute force way is impossible, many of which are not even known or understood \cite{luo2021functional,luo2023and}. To meet these challenges, 
we introduce a notion of degradation-independent latent representation (DiR) that is robust against camera-captured non-ideal real-world images and propose a new deep neural network approach to learn the DiR. 
Fig.~\ref{fig_DI_motivation} illustrates the difference between our DiR learning paradigm and the canonical learning paradigm.
Within the framework of our DiR learning, the learned DiR representations contain intrinsic information of the original degradation-free image and hence it is more general and robust with respect to unseen degradation types and severity when used in downstream computer vision and image processing tasks. The DiR learning, carried out by a deep neural network called DiRNet, is self-supervised, and the DiRNet is trained with two learning objectives: any two degraded images are mapped to the same DiR, and all DiRs are sampled from a normal distribution, which is learning a variational auto-encoder (VAE) by the multi-view mutual information (MMI) maximization\cite{tishby2000information,HwangEtalNeurIPS2020,federici2020learning}.  

Moreover, we take advantage of known priors of the ISP workflow to synthesize original and degraded pairs, so that the above baseline DiR can be reinforced by an auxiliary self-supervised learning branch. This auxiliary branch learns the reference representation $\mathbf{r}^{\ast}$ of the degradation-free images, using high-quality images (surrogate ground truth).  The baseline DiR result $\mathbf{r}^{(0)}$ is statistically aligned with the result $\mathbf{r}^{\ast}$ of the auxiliary branch by an alignment network module. This alignment module is trained using the aforementioned synthesized pairs and learns to align the baseline DiR $\mathbf{r}^{(0)}$ to the degradation-free representation reference (DfR) $\mathbf{r}^{\ast}$, yielding a refined DiR $\mathbf{r}^{+}$; finally, the refined DiR $\mathbf{r}^{+}$ is ready for being used in downstream tasks. Overall, the refined DiR benefits from
the best of both self-supervised and supervised learning, and strikes a good balance between the model's robustness and precision for applications. 

Our main contributions and key results can be summarized as follows:
\begin{itemize}
    \item A novel method is proposed for self-supervised learning of the DiR representation by multivariate mutual information maximization, which requires only degraded training images. 
    
    \item A joint learning approach is employed to refine the DiR representation for optimal performance in the targeted end-task network. 

    \item The versatility of the proposed methods across a wide range of tasks, including image restoration, object detection, and instance segmentation, is supported by ample empirical evidence. 
    
\end{itemize}

\vspace{-6pt}
\section{Related Work}


\paragraph{Degradation blind image restoration network}
There is a long history of research on image restoration tasks that are blind to degradation parameters, such as blind image deconvolution algorithms \cite{levin2009}.  Accompanying the proliferation of deep learning based image restoration methods, recent years have seen a number of papers \cite{DMMGuo2020,guo2019pipeline,wang2021real,Pan_2023_ICCV,scpgabnet,guo2023selfsupervised,li2023learning,luo2023and}  on the topic of improving the generalization ability of image restoration networks. Wang \etal \cite{wang2021real} used a high-order degradation model to mimic real-world degradations. Luo \etal \cite{luo2023and} proposed a novel adversarial neural degradation model to simulate degraded training data. Li \etal \cite{li2023learning} introduced a distortion invariant representation learning paradigm from the causality perspective, to improve the generalization ability of restoration networks for unknown degradations.

\vspace{-15pt}
\paragraph{End-to-end neural network solutions for ISP pipelines}

A more holistic design of neural networks \cite{schwartz2018deepisp,nishimura2018automatic,yoshimura2023rawgment} for image restoration is to replace the cascaded stages of the hardware ISP pipeline with a single end-to-end mapping from raw Bayer sensor data to the intended ideal RGB images, instead of postprocessing the output images of the camera ISP. 
For instance, Chen \etal \cite{chen2018learning} proposed a CNN method, called SID (see in the dark), which directly converts a low lighting noisy Bayer image to a corresponding properly exposed RGB image, without generating the underexposed ineligible RGB image first and then restoring it.
Tseng \etal \cite{Tseng19} leveraged a convolutional neural network to parameterize a differentiable mapping between the ISP configuration space and evaluation metrics of downstream tasks and validated their method on the tasks of object detection, and extreme low-light imaging. Yu \etal \cite{yu2021reconfigisp} replaced the conventional ISP with a trainable ISP framework that improves the quality of ISP-generated results by using a differential neural architecture search algorithm to search for the optimal ISP architecture and hyperparameters.

\vspace{-15pt}
\paragraph{Image enhancement for high-level tasks}
In general practice, image restoration networks tend to be optimized for human visual perception rather than for computer vision tasks. However, the human vision-centric neural network may not provide optimal results on high-level computer vision tasks. This motivates the development of a family of image restoration networks that are tuned in line with the subsequent computer vision tasks, such as recognition, detection, and classification \cite{Cui_2021_ICCV,yu2021reconfigisp,qin2023learning}. 
For example, Cui \etal \cite{Cui_2021_ICCV} proposed a multitask auto-encoding transformation model to learn the intrinsic visual structure by considering the physical sensor noise model in ISP, demonstrating effectiveness for dark object detection. 
Mosleh \etal \cite{isp_opt_cvpr20} investigated a so-called ``hardware-in-the-loop'' strategy to directly optimize
the hyperparameters of hardware ISP pipelines, by solving a nonlinear multi-objective optimization problem with a 0th-order stochastic solver. 
Qin \etal \cite{qin2023learning} 
designed a sequential CNN model to recurrently adjust the ISP hyperparameters for downstream tasks and demonstrate the advantages of the model. 



\section{Approach}

In this section, we first detail the DiRNet mentioned in the introduction to learn the baseline DiR $\mathbf{r}^{(0)}$ of multiple degraded images. The DiRNet training is driven by the MMI maximization. 
Then, we discuss how to learn the degradation-free latent representation $\mathbf{r}^{\ast}$ using a variant of DiRNet. The two latent representations $\mathbf{r}^{(0)}$ and $\mathbf{r}^{\ast}$ are prepared to train 
an alignment network to produce a refined DiR $\mathbf{r}^{+}$. Finally, we elaborate on the process to jointly optimize the alignment network and an auxiliary network for a downstream task.

\subsection{\label{sec_ss_dir} Optimizing DiRNet by MMI Maximization} 

\begin{figure}[t]
  \centering
  \includegraphics[width=1\linewidth]{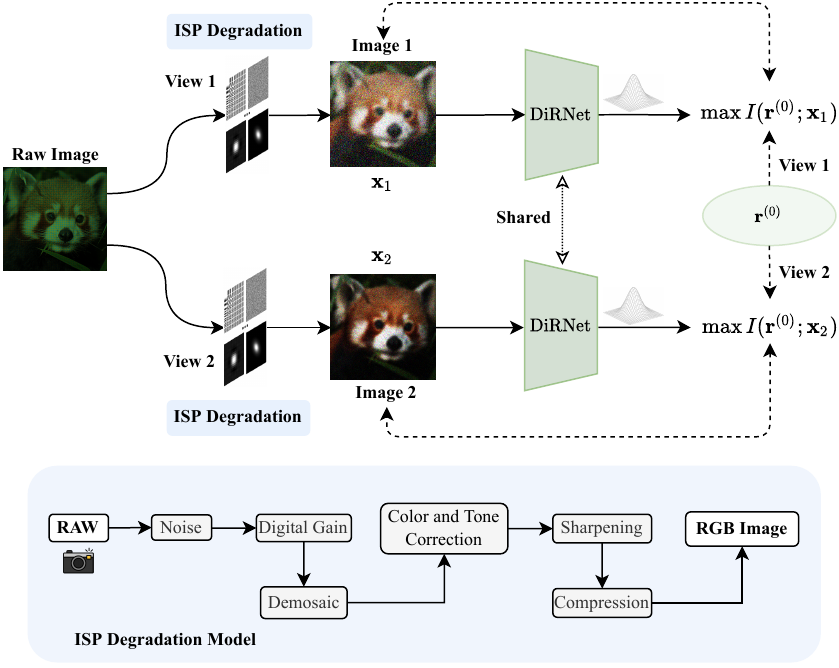}
  \caption{Self-supervised MMI maximization for DiRNet.}
  \label{ssl_self_supervised_domain}
  \vspace{-15pt}
\end{figure}

\paragraph{Mutual information maximization}
To begin, we pave the way for computing the multi-view mutual information (MMI) maximization by introducing a mutual information estimator. For estimating mutual information $I(\mathbf{x};\mathbf{y})$ of two random variables $\mathbf{x}$ and $\mathbf{y}$, it is equivalent to computing the Kullback-Leibler (KL) divergence  $\mathcal{D}_{\mathrm{KL}}(p(\mathbf{x},\mathbf{y})\,||\,p(\mathbf{x})p(\mathbf{y}))$.
However, the KL divergence has no theoretical upper bound. Maximizing it cannot guarantee the convergence of the optimization process. Following previous works \cite{nowozin2016f,van2018representation}, we exploit a non-KL divergence named Jensen-Shannon divergence (JSD) to devise the lower bound of mutual information, which satisfies the below proposition. 
\begin{proposition}
\vspace{-7pt}
Let $\mathbf{x}$ and $\mathbf{y}$ represent two random variables, 
the  $\mathbf{J} = p(\mathbf{x},\mathbf{y})$ and $\mathbf{M}=p(\mathbf{x})p(\mathbf{y})$ are the joint and the product of marginals of the two variables, respectively. 
The mutual information between the variables satisfies\footnote{\label{mi_derivation}The derivation is in Sec.B and Sec.C of the supplementary material.} 
\begin{align}
  I(\mathbf{x};\mathbf{y}) &:=  \mathcal{D}_{\mathrm{JSD}}(\mathbf{J};\mathbf{M}) \nonumber \\
  \geq& \mathbb{E}_{z\sim \mathbf{J}}\left[-\sigma(-\mathcal{F}_{\omega}(z))\right] - \mathbb{E}_{z^{\prime}\sim \mathbf{M}}\left[\sigma(\mathcal{F}_\omega(z^{\prime}))\right],
  \label{mi_jsd}
\end{align}
\label{MI_lowerbound}
\vspace{-20pt}
\end{proposition}
where $\sigma(t)=\mathrm{log}(1+e^t)$, and the discriminator function $\mathcal{F}_{\omega}$ \cite{nowozin2016f} is modeled by a neural network with parameters $\omega$.  

\vspace{-15pt}
\paragraph{Multi-view mutual information maximization}
The baseline DiR $\mathbf{r}^{(0)}$ is a degradation-independent representation shared among all degraded images that originate from an ideal degradation-free image. From the perspective of information theory, the DiR representation should satisfy the criterion of shared information maximization between it and multiple degraded observations. This shared information maximization is so-called multi-view mutual information (MMI)  maximization\cite{federici2020learning} a.k.a interaction information maximization \cite{tishby2000information} for the three variables. We state its definition as follows:
Given two degraded images $\mathbf{x}_1$ and $\mathbf{x}_2$, their shared latent DiR $\mathbf{r}^{(0)}$ can be found by maximizing their MMI  $I(\mathbf{r}^{(0)};\mathbf{x}_1;\mathbf{x}_2)$ that is defined as  
\begin{align}
  I(\mathbf{r}^{(0)};\mathbf{x}_1;\mathbf{x}_2) &=  I(\mathbf{r}^{(0)};\mathbf{x}_1) - \underbrace{I(\mathbf{r}^{(0)};\mathbf{x}_1|\mathbf{x}_2)}_{\text{Redundancy}}\label{interaction_info_def1}\\
  &= I(\mathbf{r}^{(0)};\mathbf{x}_2) - I(\mathbf{r}^{(0)};\mathbf{x}_2|\mathbf{x}_1), \label{interaction_info_def2}
\end{align}
where Eq.~(\ref{interaction_info_def2}) holds due to symmetry. The conditional mutual information $I(\mathbf{r}^{(0)};\mathbf{x}_1|\mathbf{x}_2)$ represents the incremental mutual information between  
$\mathbf{r}^{(0)}$ and the degraded observation $\mathbf{x}_1$ when given another degraded observation $\mathbf{x}_2$. This incremental information denotes the information not shared by the observations $\mathbf{x}_1$ and $\mathbf{x}_2$, due to the differences in various degradation sources, which should be minimized. Maximizing $I(\mathbf{r}^{(0)};\mathbf{x}_1;\mathbf{x}_2)$ encourages $\mathbf{r}^{(0)}$ to encode the degradation-independent information shared by the given multiple degraded images. 

The network DiRNet for learning the baseline DiR is of VAE type \cite{higgins2016beta}.
The DiRNet takes the ISP-generated images (\ie camera captured) as inputs and maps them into a canonical representation space \ie the DiR latent space, by removing the degradation-specific information.
Fig.~\ref{ssl_self_supervised_domain} shows the training process for the DiRNet. 
To continue the development, we introduce the following proposition \cite{HwangEtalNeurIPS2020}:
\begin{proposition}
  Given two degraded images $\mathbf{x}_1$ and $\mathbf{x}_2$, and the baseline DiR representation $\mathbf{r}^{(0)}$, the conditional mutual information $I(\mathbf{r}^{(0)};\mathbf{x}_1|\mathbf{x}_2)$ satisfies
  \begin{equation}
  {\fontsize{9}{0}\selectfont
    \begin{aligned}
        I(\mathbf{r}^{(0)};\mathbf{x}_1|\mathbf{x}_2)\leq \mathbb{E}_{ p(\mathbf{x}_1,\mathbf{x}_2)}[\mathcal{D}_{\mathrm{KL}}(p(\mathbf{r}^{(0)}|\mathbf{x}_1,\mathbf{x}_2)||p_\varphi(\mathbf{r}^{(0)}|\mathbf{x}_2))].
    \end{aligned}
    }
  \end{equation}
\label{CMI_lowerbound}
\vspace{-20pt}
\end{proposition}
To maximize $I(\mathbf{r}^{(0)};\mathbf{x}_1;\mathbf{x}_2)$, 
the overall training objective\footnote{The full derivation of $\mathcal{L}_{0}$ can be found in Sec.B of the supplementary material. } 
of DiRNet is defined by the average of Eq.~(\ref{interaction_info_def1}) and Eq.~(\ref{interaction_info_def2}):
\begin{align}
  \mathcal{L}_{0} =  &- \frac{I(\mathbf{r}^{(0)};\mathbf{x}_1) + I(\mathbf{r}^{(0)};\mathbf{x}_2)}{2} \nonumber \\
  &+ \lambda \mathbb{E}_{p(\mathbf{x}_1,\mathbf{x}_2)}[\mathcal{D}_{\mathrm{AKL}}]  \nonumber
  \\
  &+ \beta \mathcal{D}_{\mathrm{KL}}\left[p(\mathbf{r}^{(0)})\,||\, \mathcal{N}(0, \mathbf{I})\right],
  \label{L_DI}
\end{align}
in which we can maximize the upper bound of the conditional mutual information according to Proposition~\ref{CMI_lowerbound}. 
In Eq.(\ref{L_DI}), $\lambda$ and $\beta$ are the trade-off hyperparameters. The DiR representation is regularized by the Gaussian prior $\mathcal{N}(0, \mathbf{I})$ as in VAE \cite{higgins2016beta}. $\mathcal{D}_{\mathrm{AKL}}$ stands for the average of $\mathcal{D}_{\mathrm{KL}}(p(\mathbf{r}^{(0)}|\mathbf{x}_1,\mathbf{x}_2)\,||\,p_\varphi(\mathbf{r}^{(0)}|\mathbf{x}_1))$ and $\mathcal{D}_{\mathrm{KL}}(p(\mathbf{r}^{(0)}|\mathbf{x}_1,\mathbf{x}_2)\,||\,p_\varphi(\mathbf{r}^{(0)}|\mathbf{x}_2))$. The trainable network is the DiRNet $p_\varphi$ that produces the baseline DiR and an auxiliary network to estimate the upper bound \cite{HwangEtalNeurIPS2020}.

\subsection{\label{sec_dgfree_ref}Degradation-Free Representation Reference}

Although the baseline DiR $\mathbf{r}^{(0)}$ aims to distill the degradation-invariant intrinsic elements and structures, 
it may have biases toward the degraded data because the learning process of $\mathbf{r}^{(0)}$ has not observed degradation-free samples. To compensate for this insufficiency, we collect a large set of high-quality RGB images and assume them to be free of degradation. As these images are captured by real non-idealistic cameras the degradation-free assumption is not strictly true, but it is acceptable for practical purposes.  These degradation-free images are used to learn the degradation-free latent representation (DfR) $\mathbf{r}^{\ast}$;  the above two latent representations $\mathbf{r}^{(0)}$ and $\mathbf{r}^{\ast}$ will be aligned to produce a refined DiR $\mathbf{r}^{+}$ by a so-called alignment network, whose construction and rationale will be discussed in Sec.~\ref{sec_enhanced_dir}. 

Similar to training DiRNet for the baseline DiR $\mathbf{r}^{(0)}$, 
we propose the DfRNet to learn the DfR $\mathbf{r}^{\ast}$ from a degradation-free image (surrogate ground truth).
Since this DfR learning is with respect to a single image, the MMI training for $\mathbf{r}^{\ast}$ is simplified to the mutual information maximization \ie maximizing  $I(\mathbf{r}^{\ast};\mathbf{y}^{\ast})$, which can be solved by maximizing the lower bound derived by Proposition~\ref{MI_lowerbound}.
Moreover, following VAE \cite{higgins2016beta}, we train the decoder $D_{\ast}$ that forces the learned DfR to be able to reconstruct the degradation-free image. The objective of the above DfR learning is to minimize $\mathcal{L}_{\ast}$ defined as follows 
\begin{align}
  \mathcal{L}_{\ast} =  &-I(\mathbf{r}^{\ast};\mathbf{y}^{\ast}) 
  + \mathbb{E}\left[\left\|D_{\ast}(\mathbf{r}^{\ast}) - \mathbf{y}^{\ast}\right\|_1 \right] \nonumber \\
  & + \beta^{*} \mathcal{D}_{\mathrm{KL}}\left[p(\mathbf{r}^{\ast})\,||\, \mathcal{N}(0, \mathbf{I})\right],
  \label{L_degradation_free}
\end{align}
where $\beta^{*}$ is the trade-off hyperparameter. 
The defined training objective in Eq.~(\ref{L_degradation_free}) is the same as the InfoVAE \cite{zhao2017infovae}.

\subsection{\label{sec_enhanced_dir} Alignment Network for Refined DiR}

\paragraph{Alignment network}
As mentioned in the introduction, we can neutralize, to a certain extent, possible biases in the baseline DiR $\mathbf{r}^{(0)}$ by  
aligning $\mathbf{r}^{(0)}$ with the above DfR $\mathbf{r}^{\ast}$ in distribution.  This is accomplished by a so-called alignment network $\mathcal{A}$. 
To guide the statistical alignment of $\mathbf{r}^{(0)}$ and $\mathbf{r}^{\ast}$, 
we also feed the degraded training image x to network DfRNet and use the resulting latent representation 
$\mathop{\mathbf{r}}\limits^{\rightarrow}=\text{DfR}(\mathbf{x})$, called the pilot representation.   
This guided alignment process is a mapping function 
\begin{equation}
\resizebox{.3\textwidth}{!}{%
$\begin{aligned}
\mathcal{A}(\mathbf{r}^{(0)},\mathop{\mathbf{r}}\limits^{\rightarrow}) : 
\mathbf{r}^{(0)} 
\xrightarrow[\mathop{\mathbf{r}}\limits^{\rightarrow} = \text{DfR}(\mathbf{x})]{\text{Guided by } \mathop{\mathbf{r}}\limits^{\rightarrow} } \mathbf{r}^{+}.
\label{guide_rmn}
\end{aligned}$
}
\end{equation}

Next, we describe the architecture of the alignment network $\mathcal{A}$ and discuss the role of the pilot representation $\mathop{\mathbf{r}}\limits^{\rightarrow}$. 

\paragraph{Network structure} 
\begin{figure}[htbp]
\vspace{-21pt}
\centering
\includegraphics[width=1\linewidth]{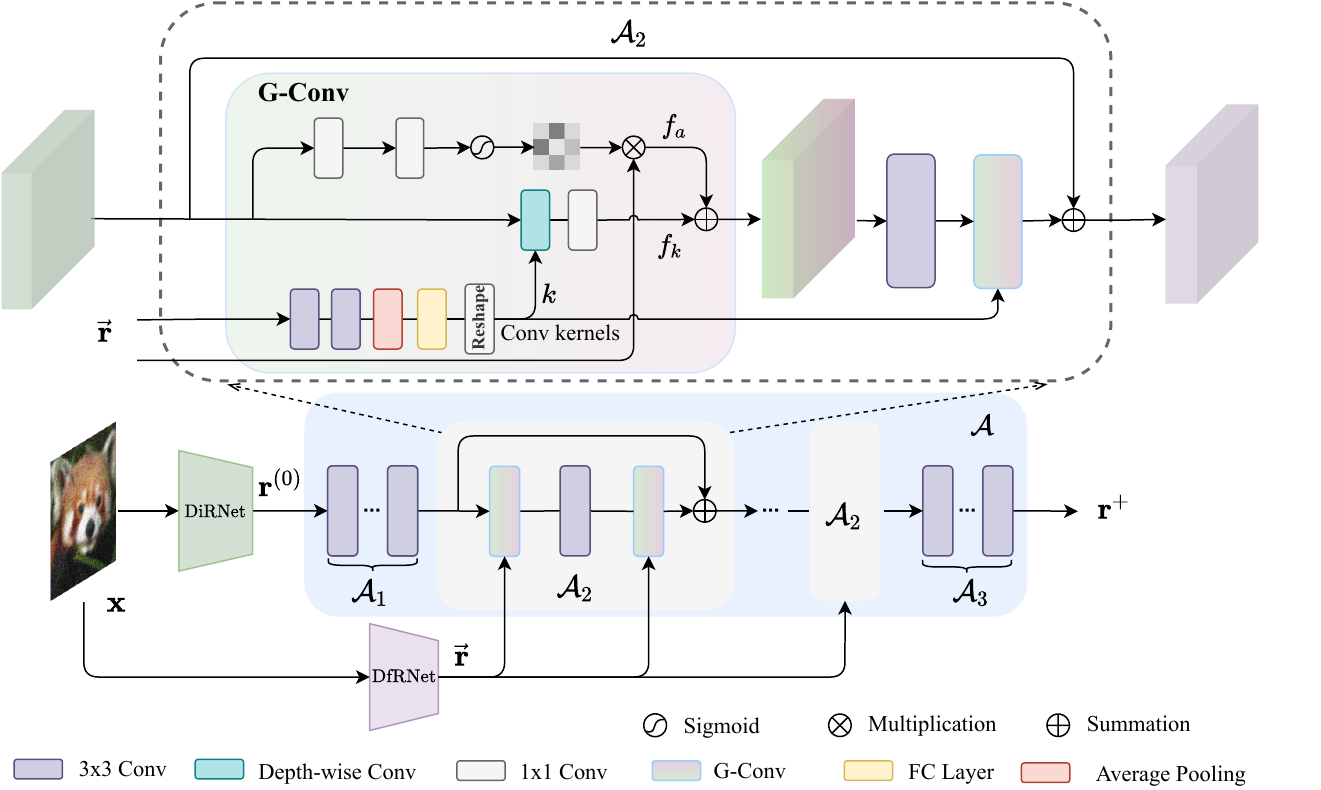}
\vspace{-17pt}
\caption{Illustration of the alignment network.}
\label{fig:tga_network}
\vspace{-10pt}
\end{figure}
Fig.~\ref{fig:tga_network} shows the network structure of the alignment network $\mathcal{A}$. It consists of three cascaded modules $\mathcal{A}_1$, $\mathcal{A}_2$ and $\mathcal{A}_3$. The front module $\mathcal{A}_1$ of $m_1$ convolution layers is used to extract the features from the input image. 
The middle module $\mathcal{A}_2$ of $m_2$ layers is designed to modulate the latent features with the guidance of the pilot DfR $\mathop{\mathbf{r}}\limits^{\rightarrow}$. 
The end module $\mathcal{A}_3$ of $m_3$ convolution layers is used to produce the final refined DiR $\mathbf{r}^{+}$. 

In the guidance module $\mathcal{A}_2$, $\mathop{\mathbf{r}}\limits^{\rightarrow}$ is fed to convolution layers with $3\times 3$ kernels, followed by the average pooling layer to
obtain the modulation feature tensors. These modulation tensors are then reshaped to form $k$ adaptive convolution kernels. We use a so-called G-Conv module formed by these adaptive convolution kernels to convolve with the input features to produce the guided features $f_k$ which are embedded with the information of the pilot DfR $\mathop{\mathbf{r}}\limits^{\rightarrow}$. 
In addition, we exploit a spatial attention branch to assist the guidance information extraction. This attention branch predicts the attention maps to strengthen the feature units that have a high correlation to pilot DfR $\mathop{\mathbf{r}}\limits^{\rightarrow}$, yielding the attention-weighted intermediate features $f_a$.  
Finally, the guided features $f_k$ and the attention-weighted features $f_a$ are fused by an addition layer to produce the refined DiR $\mathbf{r}^{+}$, which is ready for use in downstream tasks. In our experiments, we {\em jointly} train this alignment network $\mathcal{A}$ and a task-aligned auxiliary network $\mathcal{T}$, which will be introduced in Sec.~\ref{sec_training_obj}.

\begin{figure}[htbp]
\vspace{-10pt}
  \centering
  \includegraphics[width=0.7\linewidth]{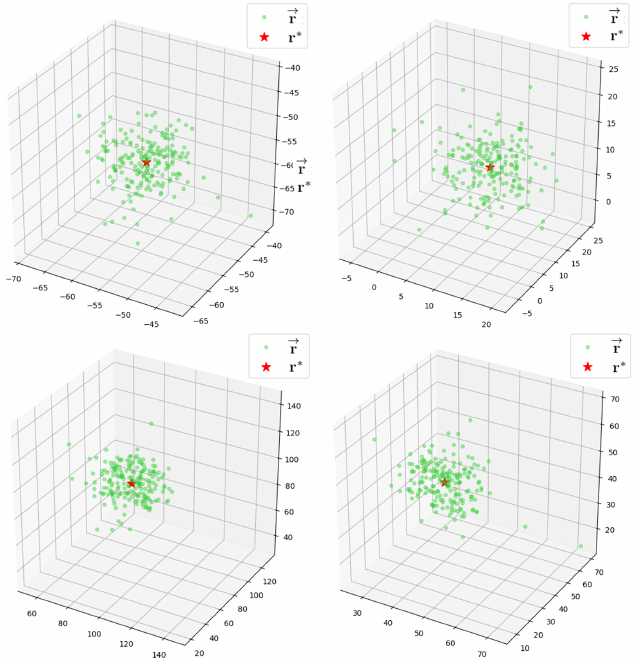}
  \vspace{-10pt}
  \caption{Visualization of the pilot DfR representations $\mathop{\mathbf{r}}\limits^{\rightarrow}$ and the expected DfR representations $\mathbf{r}^{\ast}$. The information provided by the pilot DfR can guide the alignment to find the optimal DfR.}
  \label{fig:guidance_rational}
\vspace{-20pt}
\end{figure}

\paragraph{Discussion on the pilot DfR $\mathop{\mathbf{r}}\limits^{\rightarrow}$} 
To explain the rationale behind the guidance design of the alignment network, we conduct experiments for visualization. More specifically, 
we choose four different degradation-free images to extract their corresponding DfRs $\mathbf{r}^{\ast}$.
We then use the ISP pipeline as in \cite{qin2023learning} by adding noise to generate 200 various degraded copies of each degradation-free image. 
These degraded samples, which share the same image content but have different ISP degradations, are fed into the network $\text{DfR}$ to produce the pilot DfRs $\mathbf{r}^{\ast}$. 
Fig.~\ref{fig:guidance_rational} shows the visualized results by using t-SNE \cite{van2008visualizing}, where the red stars denote the expected ground truth DfRs $\mathbf{r}^{\ast}$ for the alignment network training, and the green stars stand for the pilot DfRs $\mathop{\mathbf{r}}\limits^{\rightarrow}$. 
We can observe that the green points are projected near the red points, suggesting that the pilot DfRs have a strong correlation with the ground truth DfRs. This is because the pilot DfRs and the ground truth DfRs share the same degradation-free content. 
Inspired by this phenomenon, the alignment network is designed to take into account the assistant information provided by $\mathop{\mathbf{r}}\limits^{\rightarrow}$ to guide the refinement of the baseline DiR.

\subsection{\label{sec_training_obj} Joint Task-Aligned Refinement Learning}

\begin{figure}[htbp]
  \centering
  \includegraphics[width=1\linewidth]{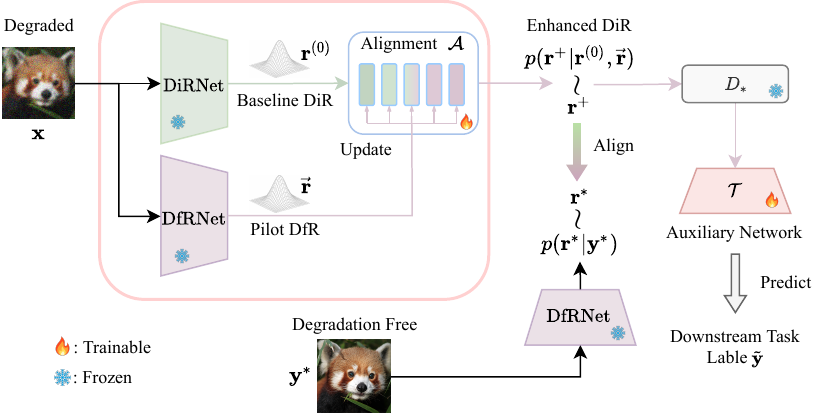}
  \vspace{-17pt}
  \caption{Illustration of the joint task-aligned refinement learning.}
  \label{fig:guided_alignment}
\vspace{-14pt}
\end{figure}

For training the alignment network, we simulate the Bayer RAW data from the degradation-free images $\mathbf{y}^{\ast}$ (surrogate ground truth) by using the inverse ISP \cite{kimPAMIISP}.  These RAW data are used to generate the degraded RGB images $\mathbf{x}$ with the ISP software tool \cite{qin2023learning}. Fig.~\ref{fig:guided_alignment} illustrates the joint task-aligned refinement learning process.
The DiRNet and DfRNet are frozen to extract the paired representations $(\mathbf{r}^{(0)},\mathbf{r}^{\ast})$ from the paired degraded and degradation-free images $(\mathbf{x},\mathbf{y}^{\ast})$. 
 The alignment network is driven by the pairs $(\mathbf{r}^{(0)},\mathbf{r}^{\ast})$ to learn the mapping function $\mathcal{A}:$ $\mathbf{r}^{(0)} \rightarrow \mathbf{r}^{+}$, where the final refined DiR $\mathbf{r}^{+}$ lies in the degradation-free latent representation space determined by the DfRNet. 

The paired images $(\mathbf{x}, \mathbf{y}^{\ast})$ and the associated labels $\tilde{\mathbf{y}}$ w.r.t. the downstream task are used to supervise the joint training of the alignment network $\mathcal{A}$ and the task-related auxiliary network $\mathcal{T}$.
As the auxiliary network for the downstream task usually takes an RGB image as input, we adopt the decoder $D_{\ast}$ trained for the DfRNet to reconstruct the RGB images from the refined DiRs $\mathbf{r}^{+}$.  
The joint training loss in the alignment phase is to minimize $\mathcal{L}_{+}$ defined as 
\begin{align}
  \mathcal{L}_{+} = & \mathbb{E}\left[\| \mathbf{r}^{+} - \mathbf{r}^{\ast} \|_1\right] + \gamma_1 \mathcal{L}_{\text{task}}(\mathcal{T}(D_{\ast}(\mathbf{r}^{+})),\tilde{\mathbf{y}}) \nonumber\\
  &+ \gamma_2 \mathbb{E}\left[\| D_{\ast}(\mathbf{r}^{+}) -\mathbf{y}^{\ast}\|_1\right], 
  \label{loss_T}
\end{align}
where $\gamma_1$ and $\gamma_2$ are the trade-off parameters, the  DfR $\mathbf{r}^{\ast} = \text{DfR}(\mathbf{y}^{\ast})$, and the refined DiR representation $\mathbf{r}^{+} = \mathcal{A}(\mathbf{r}^{(0)}, \mathop{\mathbf{r}}\limits^{\rightarrow})$, where $\mathbf{r}^{(0)}=\text{DiR}(\mathbf{x})$ and $\mathop{\mathbf{r}}\limits^{\rightarrow}=\text{DfR}(\mathbf{x})$. 
The alignment network $\mathcal{A}$ and the auxiliary network $\mathcal{T}$ are jointly optimized, whereas the pretrained networks DiRNet and DfRNet as well as the decoder $D_{\ast}$ are frozen. The second term $\mathcal{L}_{\text{task}}$ is a task-related loss for a downstream task, such as the cross-entropy for image segmentation.
The third term is an image reconstruction loss that forces the refined DiR to inherit the complete information from the degradation-free image through the pixel-wise reconstruction restriction. We summarize the DiR learning process in Algorithm \ref{alg:DiR}.

\begin{algorithm}
  \caption{DiR Learning and Task-Aligned Refinement}
  \label{alg:DiR}
  \setlength{\tabcolsep}{1pt}
    \fontsize{8}{9}\selectfont
  \begin{algorithmic}[1]
    \Require{ Degradation-free RAW image $\mathbf{y}$ and corresponding degradation-free RGB image $\mathbf{y}^{\ast}$, task-related  label $\tilde{\mathbf{y}}$}
    \Statex
    \textbf{Stage~\uppercase\expandafter{\romannumeral1}:} \Repeat
    \State Generate two images $\mathbf{x}_1$ and $\mathbf{x}_2$ from a degradation-free image $\mathbf{y}$ using two randomly sampled ISP degradations. 
    \State Train DiRNet using $(\mathbf{x}_1,\mathbf{x}_2)$ ($\min \mathcal{L}_{0}$ in Eq.~(\ref{L_DI}))
    \State Train DfRNet using $\mathbf{y}^{\ast}$ ($\min \mathcal{L}_{\ast}$ in Eq.~(\ref{L_degradation_free})) 
    \Until{reaching the maximum number of epochs}
    \Statex \textbf{Stage~\uppercase\expandafter{\romannumeral2}:}
    \Repeat
    \State Generate an image $\mathbf{x}$ from a degradation-free image $\mathbf{y}$ using randomly sampled ISP degradations. 
    \State  $\mathop{\mathbf{r}}\limits^{\rightarrow}=\text{DfR}(\mathbf{x})$,\, $\mathbf{r}^{\ast}=\text{DfR}(\mathbf{y}^{\ast})$
    \State $\mathbf{r}^{+}=\mathcal{A}(\mathbf{r}^{(0)},\mathop{\mathbf{r}}\limits^{\rightarrow})$
    \State Train the alignment network $\mathcal{A}$ using the paired $(\mathbf{x},\mathbf{r}^{\ast})$ with the guidance of the pilot DfR $\mathop{\mathbf{r}}\limits^{\rightarrow}$, and jointly train the auxiliary  $\mathcal{T}$ using the pairs of $(D_{\ast}(\mathbf{r}^{+}),\tilde{\mathbf{y}})$ ($\min \mathcal{L}_{+}$ in Eq.~(\ref{loss_T}))
    \Until{reaching the maximum number of epochs}
  \end{algorithmic}
\end{algorithm}

\section{Experiments}

\subsection{Implementation Details}
\paragraph{Datasets and evaluation}
As previously introduced, DiR learning comprises two stages: Stage~\uppercase\expandafter{\romannumeral1}, dedicated to training the DiRNet and DfRNet, and Stage~\uppercase\expandafter{\romannumeral2}, where the 
alignment network $\mathcal{A}$ and the task-related auxiliary network $\mathcal{T}$ are jointly trained. For the training dataset, we adopt the DIV2K dataset \cite{Agustsson_2017_CVPR_Workshops} and degrade the images by synthetic ISP degradations. To simulate ISP degradations, we begin by converting RGB images into RAW data by applying the reversed ISP model \cite{kimPAMIISP}. Subsequently, we employ a synthetic ISP pipeline \cite{isp_opt_cvpr20, qin2023learning} with random parameters to generate degraded ISP-generated images from the simulated RAW data\footnote{Additional details can be found in the supplementary material}. 
In Stage~\uppercase\expandafter{\romannumeral2}, we employ the identical simulation process used in Stage~\uppercase\expandafter{\romannumeral1} to synthesize ISP-degraded images and use the  MS COCO dataset \cite{lin2014microsoft} for experiments.  

For evaluation, we test the proposed method on three downstream tasks, including 1) the image restoration experiments on real-world datasets including the PolyU-Real \cite{xu2018realworld}, NC12 \cite{nc2015}, Nam \cite{nam2016holistic}, and DND\footnote{We randomly select 500 images for the experiments} \cite{pltz2017benchmarking} datasets, where the test images involve real complex ISP degradations, 2) the object detection experiments on the MS COCO \cite{lin2014microsoft} dataset, and 3) the image segmentation experiments on the MS COCO dataset and the ADE20k dataset \cite{zhou2017ade}.

\vspace{-13pt}
\paragraph{Training settings}

At Stage~\uppercase\expandafter{\romannumeral1}, we randomly select the contrastive degraded pairs to enable the MMI maximization for the training of the DiRNet. For training the DfRNet, we follow the same setting as in \cite{zhao2017infovae}, by setting the parameter $\beta^{\ast}=1$ in Eq.~(\ref{L_degradation_free}). We adopt the encoder network of \cite{isola2017image} as the architectures of the DiRNet and the DfRNet. During training, 
the learning rate is
initially set to $10^{-4}$ and reduced to $10^{-6}$ after 200 epochs. 
At Stage~\uppercase\expandafter{\romannumeral2}, we randomly form pairs of degraded and degradation-free RGB images for training the alignment network. The learning rate is initially set to $10^{-3}$ and reduced to $10^{-6}$ after 300 epochs. For the image restoration task, we set $\gamma_1=0,\gamma_2=1$, because the decoder $D_{\ast}$ can serve as the auxiliary network for the RGB image reconstruction using the refined DiR $\mathbf{r}^{+}$. For the object detection and image segmentation tasks, the trade-off parameters are set to $\gamma_1=2,\gamma_2=1$ and $\gamma_1=5,\gamma_2=1$, respectively. The hyperparameters $m_1, m_2$, and $m_3$ in the alignment network are separately set to 4, 2, and 4. 

\vspace{-5pt}
\subsection{Results on ISP-Degraded Image Restoration}

First, we present the experimental results of the proposed method and compare them with those of eight competing methods \cite{dabov2007image, du2020learning,Guo2019Cbdnet,Zamir2021MPRNet,Wang_2022_CVPR,li2023learning,Pan_2023_ICCV},
on the restoration of ISP-degraded images.  The task is to remove ISP artifacts caused by complex coupling of multiple degradation sources, including sensor noises, demosaicing noises, compression noises, and possibly adverse effects of incorrect settings of ISO and gamma correction.  Here, like in all other image restoration methods for human viewers, our goal is to pursue the highest possible perceptual image quality.



\begin{figure*}[htbp]
  \centering
  \includegraphics[width=1\linewidth]{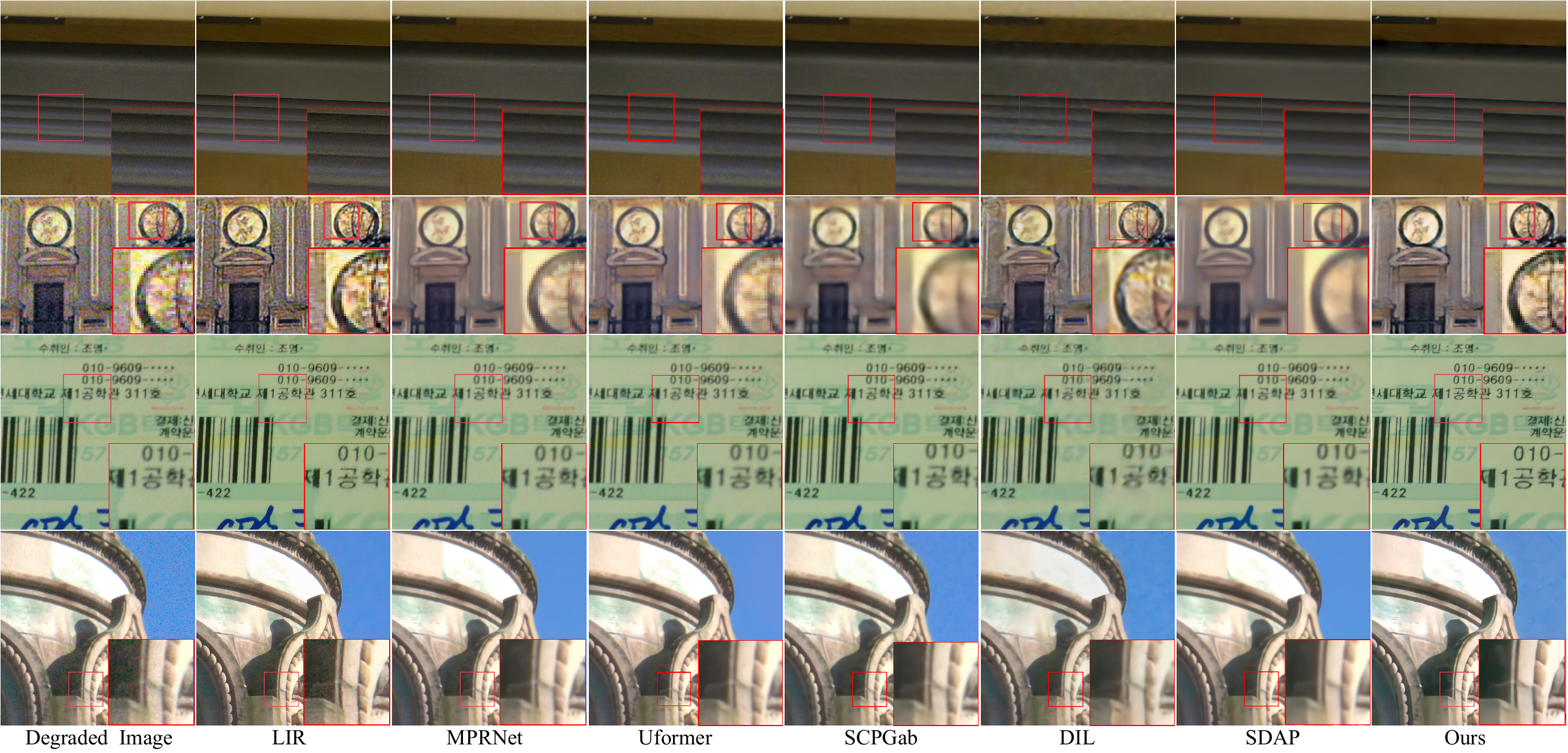}
  \vspace{-20pt}
  \caption{Qualitative comparison on real-world ISP-degraded images. The degraded images, from top to bottom, are from the real-world datasets PolyU-Real \cite{xu2018realworld}, NC12 \cite{nc2015}, Nam \cite{nam2016holistic}, and DND \cite{pltz2017benchmarking}. Please zoom in for better visualization.}
  \label{real_denoise}
\vspace{-13pt}
\end{figure*}

In Fig.~\ref{real_denoise} the reader can compare the visual quality of the DiR learning based restoration method against those of SOTA image restoration methods \cite{du2020learning,Guo2019Cbdnet,Zamir2021MPRNet,Wang_2022_CVPR,scpgabnet,li2023learning,Pan_2023_ICCV},
on restoring real-world ISP-degraded images taken by four different cameras, whose true degradation models are unknown.  
We can observe that the proposed method succeeds in achieving its design goal of blind restoration, as it removes most of the ISP artifacts caused by unknown degradation sources.
On the contrary, the competing methods fail to remove complex artifacts while preserving the sharpness and clarity of edges and textures.
Table~\ref{table_real_denoise} reports the quantitative performance evaluation results. 
Since there is no ideal degradation-free ground truth, we adopt the non-reference image quality assessment methods, including NIQE \cite{Anish2013NIQE}, NRQM \cite{ma2016learning}, and BIQA \cite{Su_2020_CVPR}. 
In agreement with the visual comparisons in Fig.~\ref{real_denoise}, the proposed DiR method also outperforms the competitors in terms of all three metrics. 
Both subjective and objective evaluations demonstrate superior robustness 
of the proposed method when being applied to data unseen in the training, indicating that the DiR learning approach can distill degradation-free representations from ISP-degraded images, regardless of whether observed or not.

\begin{table}[htbp]
  \centering
 \setlength{\tabcolsep}{2pt}
 \fontsize{9}{12}\selectfont
  \scalebox{0.8}{
  \begin{tabular}{c|c|c|c|c}
     \toprule
     \multirow{2}{*}{Method} &\multicolumn{4}{c}{ NIQE $\downarrow$ / NRQM $\uparrow$ / BIQA $\uparrow$}\\
      & PolyU-Real & NC12 & Nam& DND  \\
     \midrule
     BM3D \cite{dabov2007image}  &  9.72/2.60/40.5 &  12.78/3.75/30.8 & 12.58/2.56/41.2& 12.89/1.84/40.1 \\
     CBDNet \cite{Guo2019Cbdnet}  & 7.32/3.78/44.1 & 10.67/6.88/33.1 & 10.26/3.62/42.1 & 9.74/2.96/41.1 \\
      LIR \cite{du2020learning} & 8.37/4.52/48.1 & 11.41/7.19/33.7  & 10.69/4.25/49.8 & 11.86/3.19/44.3 \\
     MPRNet \cite{Zamir2021MPRNet} & \textbf{6.12}/4.23/38.7 & 11.21/6.30/32.9 & \textbf{7.25}/3.94/33.9 & 9.53/2.77/45.2 \\
      Uformer \cite{Wang_2022_CVPR} & 6.25/3.96/37.4 & 10.54/6.05/31.2 & 7.92/3.62/34.6 & 9.58/2.72/45.1 \\ 
      SCPGab \cite{scpgabnet} &  8.71/3.25/41.1 & 11.75/3.93/31.9 & 10.60/3.25/40.7 &  9.80/2.53 /42.0 \\
      DIL \cite{li2023learning} & 7.28/4.10/41.8  & 10.88/6.52/30.8 & 9.11/3.59/40.8 &  \textbf{7.86}/3.72/44.7  \\
      SDAP \cite{Pan_2023_ICCV} & 9.62/3.00/46.1 &11.34/3.88/32.5 & 12.38/2.95/45.7 & 10.74/2.33/45.8 \\
     \textbf{Ours} & 6.57/\textbf{4.91}/\textbf{51.9} & \textbf{9.06}/\textbf{7.41}/\textbf{34.4} & 8.96/\textbf{4.64}/\textbf{53.6} &8.62/\textbf{3.59}/\textbf{46.9} \\
     \bottomrule
  \end{tabular}
  }
  \vspace{-5pt}
    \caption{Quantitative results on real-world ISP-degraded images.}
  \label{table_real_denoise}
\vspace{-15pt}
\end{table}

\subsection{Results on Object Detection}
\begin{figure*}[t]
  \centering
  \includegraphics[width=0.95\textwidth]{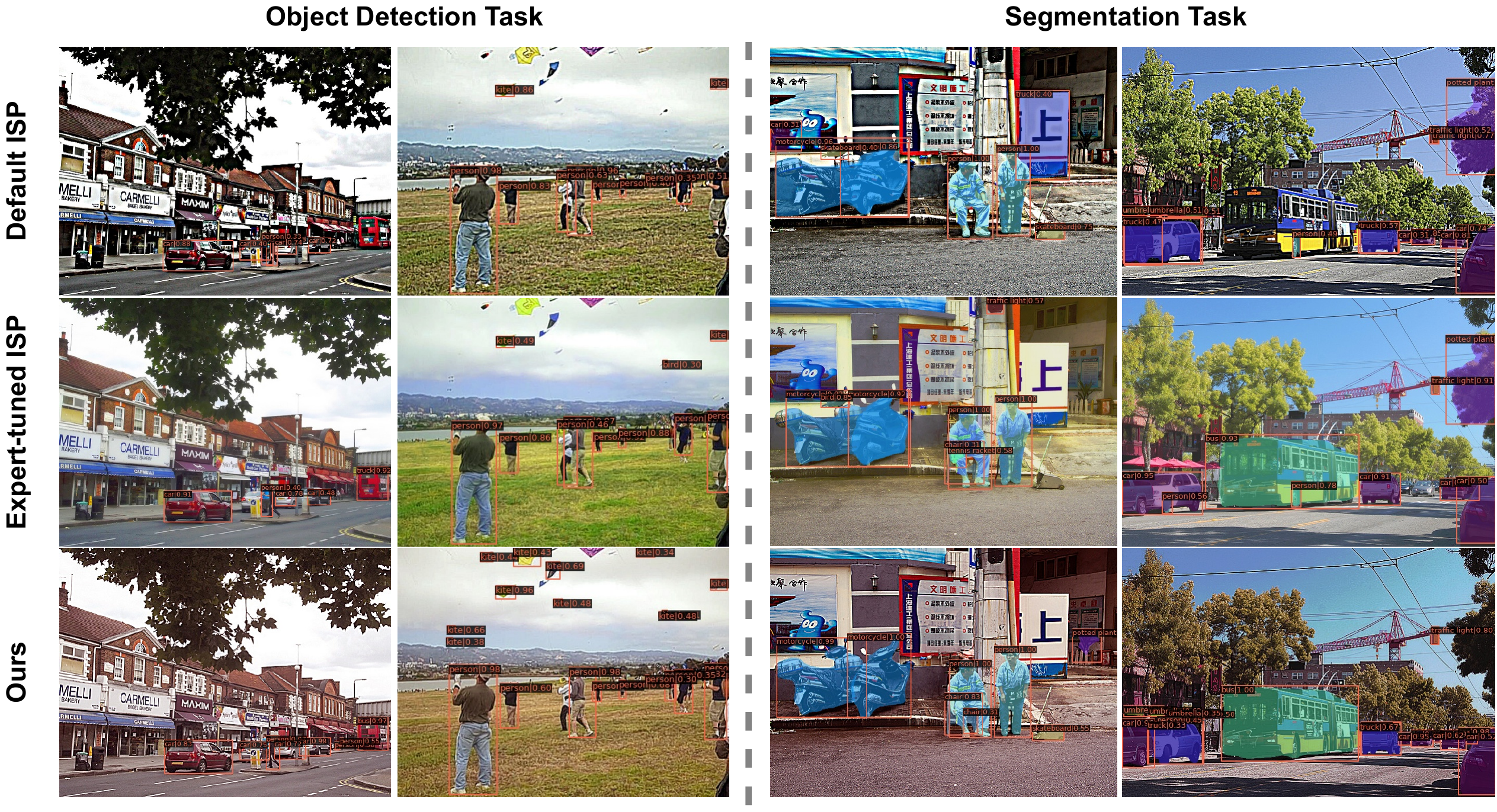}
  \vspace{-7pt}
  \caption{ We evaluate different ISP-generated images for the tasks of object detection (left) and instance segmentation (right) on the COCO dataset. From top to bottom, we show the results from the images generated by default ISP hyperparameters, expert-tuned ISP hyperparameters, and the refined DiR with the proposed method. The proposed method achieves better performance for downstream tasks.
}
\label{fig_isp_seg_detect}
\vspace{-13pt}
\end{figure*}

Following \cite{qin2023learning}, we utilize the MS COCO dataset \cite{lin2014microsoft} to evaluate our proposed method's performance on object detection. We compare our method with the SOTA neural ISP solvers \cite{qin2022attention, qin2023learning}, which replace the conventional ISP pipeline with end-to-end trained neural networks while considering the downstream task as their optimization objectives.  
Following the approach in \cite{isp_opt_cvpr20, qin2023learning}, we choose the ISP pipeline with default parameters as the baseline. Our method utilizes these ISP-generated results with default settings as inputs. For comparison, we adopt the pre-trained object detection model \cite{redmon2018yolov3} for performance benchmarking, and we assess the results using the mean accuracy score (mAP) as the evaluation metric. The quantitative results are reported in Table~\ref{tabObjdetection}. Our method demonstrates a significant improvement compared to the baseline and outperforms the SOTA solutions \cite{qin2022attention, qin2023learning}. The left side of Fig~\ref{fig_isp_seg_detect} illustrates the qualitative results. As a reference, we include the results produced by the expert-tuned ISP for visual comparison. We can observe that while the expert-tuned ISP can produce more visually appealing results aligned with human perception preferences, it is inferior to our method on the task of object detection.


\subsection{Results on Image Segmentation}

Similar to the object detection experiments, we adopt the MS COCO dataset \cite{lin2014microsoft} to evaluate the performance of image segmentation. 
We benchmark the performance using a pre-trained segmentation model \cite{he2017mask} and report the results in Table~\ref{tabSegmentCoco}. As shown, our method outperforms the SOTA method Attention-aware \cite{qin2022attention} by about 3.9\% on the mAP@0.5. 
Quantitative results are shown on the right side of Fig~\ref{fig_isp_seg_detect}. These results are in line with the observations in the object detection experiments. Our method consistently outperforms the expert-tuned ISP in achieving better segmentation results.

\begin{table}[H]
  \centering
    \fontsize{8}{9}\selectfont
  \scalebox{0.95}{
    \begin{tabular}{c|c| c c c}
        \toprule
        \multirow{2}{*}{\textbf{Methods}}& \multirow{2}{*}{ISP Model} & \multicolumn{3}{c}{mAP$\uparrow$}\\
        & & 0.5 & 0.75 & 0.5:0.95 \\
        \midrule
        Default Params  & \multirow{3}{*}{ISP \cite{isp_opt_cvpr20}} & 15.  & -& -	\\
        Blockwise-tuned \cite{nishimura2018automatic} && 20. & -& -	\\
        Hardware-tuned \cite{isp_opt_cvpr20}  && 39. & - & - \\
        \midrule
        Default Params & \multirow{4}{*}{ISP \cite{qin2023learning}} & 34.1 & 22.4 & 21.4\\
        Attention-aware \cite{qin2022attention} &  & 61.0 & 44.1 & 41.0 	\\
        Sequential-tuned \cite{qin2023learning} & & 62.8 & 45.2 & 42.3	\\
        \textbf{Ours} & & \textbf{63.9} & \textbf{46.4} & \textbf{43.3}\\
        \bottomrule
    \end{tabular}
}
\vspace{-8pt}
\captionof{table}{Quantitative results on object detection.}
  \label{tabObjdetection}
\vspace{-20pt}
\end{table}
 
\begin{table}[H]
  \centering
    \fontsize{8}{9}\selectfont
  \scalebox{0.95}{
    \begin{tabular}{c|c| c c c}
        \toprule
        \multirow{2}{*}{\textbf{Methods}}& \multirow{2}{*}{ISP Model} & \multicolumn{3}{c}{mAP$\uparrow$}\\
        & & 0.5 & 0.75 & 0.5:0.95 \\
        \midrule
        Default Params  & \multirow{2}{*}{ISP \cite{isp_opt_cvpr20}} & 12.  & -& -	\\
        Hardware-tuned \cite{isp_opt_cvpr20}  && 32. & - & - \\
        \midrule
        Default Params & \multirow{4}{*}{ISP \cite{qin2023learning}} & 22.7 & 13.2 & 12.0\\
        Attention-aware \cite{qin2022attention} &  & 52.3 & 33.4 & 31.5 	\\
        Sequential-tuned \cite{qin2023learning} & & 54.2 & 35.1 & 32.6	\\
        \textbf{Ours} & & \textbf{56.3} & \textbf{37.1} & \textbf{34.8}\\
        \bottomrule
    \end{tabular}
}
\vspace{-8pt}
\captionof{table}{Quantitative results on image segmentation.}
\label{tabSegmentCoco}
\vspace{-10pt}
\end{table}

Additionally, we conduct experiments to evaluate the performance of the proposed method in low signal-to-noise ratio situations, such as dark environments where the noise level is very high. 
To mimic the degradations, we increase the noise level in the aforementioned synthetic ISP pipeline by adding Gaussian noise with the standard deviation $\sigma$ randomly sampled between the range of $[0.15,0.35]$ and the Poisson noise with the magnitude $\lambda$ randomly sampled between the range of $[0.02,0.04]$. We compare the proposed DiR learning based method with the SOTA methods \cite{ignatov2020replacing, jincvpr23dnf} for ISP degradation removal and present the results in Table~\ref{tabSegmentADE20k}. To evaluate performance, we use two pretrained segmentation models, including DeeplabV3 \cite{chen2017rethinking} and SegFormer \cite{xie2021segformer}. We observe that our method outperforms all the competitors. For example, compared to the DNF \cite{jincvpr23dnf}, our method increases the mAcc by 7.6\% and 6.7\% when using the DeeplabV3 and SegFormer models, respectively.

\begin{table}[t]
  \centering
    \fontsize{8}{9}\selectfont
  \scalebox{0.9}{
    \begin{tabular}{c|c c c| c c c}
        \toprule
        \multirow{2}{*}{\textbf{Methods}}&  \multicolumn{3}{c|}{DeeplabV3 \cite{chen2017rethinking}} & \multicolumn{3}{c}{SegFormer  \cite{xie2021segformer}}\\ 
        & aAcc$\uparrow$&mIoU$\uparrow$&\multicolumn{1}{c|}{mAcc$\uparrow$} & aAcc$\uparrow$&mIoU$\uparrow$& mAcc$\uparrow$\\
        \midrule
        Default ISP &64.8&22.6&28.9 &74.8&36.9&45.1 \\
        PyNET \cite{ignatov2020replacing}  &67.7&25.1&31.5 &74.3&37.4&45.8\\
        DNF \cite{jincvpr23dnf} & 71.4 &29.6 & 36.9 & 75.4 &37.8 & 46.2\\
        \textbf{Ours}  & \textbf{75.5}&\textbf{34.6}&\textbf{44.5} &\textbf{79.7} &\textbf{42.4}&\textbf{52.9}\\
        \bottomrule
    \end{tabular}
}
\vspace{-7pt}
\captionof{table}{Results of image segmentation on the ADE20k \cite{zhou2017ade}.}
\label{tabSegmentADE20k}
\vspace{-19pt}
\end{table}

\subsection{Ablation Study}
Ablation studies are conducted to assess the effectiveness of the proposed components: the baseline DiR $\mathbf{r}^{(0)}$, the alignment network $\mathcal{A}$, and the pilot DfR $\mathop{\mathbf{r}}\limits^{\rightarrow}$. 
For experiments, we apply the synthetic ISP process as described in \cite{qin2023learning} to generate ISP-degraded test images using the Set14 and BSD100 \cite{MartinFTM01} datasets. 
To simulate degradation, we introduce noise into the synthetic ISP process. The added noise includes random Gaussian noise with a standard deviation ranging from $0.05$ to $0.1$ and JPEG compression noise with a quality factor randomly chosen in the range $[10,30]$. 
As a reference, we train an auto-encoder model, labeled as ``Auto-Encoder''.
This model adopts the same encoder architecture as DiRNet and the same decoder architecture as $D_{\ast}$.
The training process of this auto-encoder model follows the canonical paradigm, \ie, it learns a direct mapping from degraded images to their degradation-free counterparts.
Table~\ref{table_ablation} shows comparisons: when excluding the pilot DfR (2nd row) from the complete DiR learning technique (3rd row), we observe a decrease in PSNR by 0.2 dB and 0.4 dB for the Set14 and BSD100 datasets. In comparison to using only the baseline DiR, the inclusion of the alignment network $\mathcal{A}$ leads to a PSNR improvement from 22.4 dB to 23.9 dB and 23.8 dB to 24.6 dB on the two datasets. This enhancement is attributed to the alignment network's ability to refine the baseline DiR by aligning it with the degradation-free representation.

\begin{table}[htbp]
\vspace{-7pt}
     \centering
    \fontsize{8}{9}\selectfont
     \scalebox{0.95}{
     \begin{tabular}{ccc|c|c|c|c}
     \toprule
     \multicolumn{3}{c|}{\multirow{1}{*}{\textbf{Method}}} & \multicolumn{2}{c|}{{\bfseries Set14}} &  \multicolumn{2}{c}{{\bfseries BSD100}}\\
     \cline{4-7}
     $\mathbf{r}^{(0)}$ & $\mathcal{A}$ & $\mathop{\mathbf{r}}\limits^{\rightarrow}$ & PSNR $\uparrow$ & LPIPS $\downarrow$ & PSNR $\uparrow$& LPIPS $\downarrow$ \\
     \midrule
     \cmark& \xmark&\xmark &22.4& 0.359& 23.8 &0.379 \\
     \cmark& \cmark&\xmark & 23.9  & 0.243 & 24.6  & 0.313 \\
     \cmark& \cmark&\cmark &{\bf 24.1} & {\bf 0.228} & {\bf 25.0} & {\bf 0.288}
     \\
     \midrule
     \multicolumn{3}{c|}{Auto-Encoder} & 21.6& 0.378& 23.1 &0.408\\
      \bottomrule
     \end{tabular}
     }
     \vspace{-7pt}
      \captionof{table}{Ablation studies on each module in DiR learning. 
     }
     \label{table_ablation}
\vspace{-20pt}
\end{table}




\section{Conclusion}
This paper presents a novel method of learning degradation-independent representations for ISP-degraded images. 
In the proposed method, we devise a DiRNet to learn baseline DiR representations from diverse ISP-degraded observations in a self-supervised learning manner, ensuring adaptability to various degradation scenarios. Furthermore, we employ a guided alignment network to refine the baseline DiRs by aligning them with latent degradation-free representations. Our proposed approach demonstrates superior performance and scalability, as validated by extensive empirical results from synthetic and real-world datasets across various downstream applications.


{\small
\bibliographystyle{ieee_fullname}
\bibliography{egbib}
}

\end{document}